\documentclass[a4paper,10pt]{llncs}
\usepackage[utf8]{inputenc}
\usepackage{amsmath, amssymb}
\usepackage{graphicx}
\usepackage{xspace}
\usepackage{tikz}
\usetikzlibrary{calc}
\usepackage{todonotes}
\usepackage[defblank]{paralist}
\usepackage{url}
\usepackage[hidelinks]{hyperref}
\usepackage{comment}
\title{CoCoMoT: Conformance Checking of Multi-Perspective Processes via SMT \\(Extended Version)}
\titlerunning{CoCoMoT: Conformance Checking via SMT (Extended Version)}
\author{Paolo Felli\inst{1} \and
Alessandro Gianola\inst{1} \and
Marco Montali\inst{1} \and \\
 Andrey Rivkin\inst{1} \and Sarah Winkler\inst{1}}
\authorrunning{Felli, Gianola, Montali, Rivkin, Winkler}
\institute{%
 Free University of Bozen-Bolzano, Bolzano, Italy
\email{\{pfelli,gianola,montali,rivkin,winkler\}@inf.unibz.it}
}

\newcommand{\seq}[2][n]{{#2_1},\dots,{#2_{#1}}}
\newcommand{\dist}{\delta}
\newcommand{\set}[1]{\{#1\}}                      
\renewcommand{\vec}[1]{\mathbf{#1}} 
\newcommand{\Dom}{\mathcal D} 
\newcommand{\pre}[1]{{{}^\bullet{#1}}} 
\newcommand{\post}[1]{{{#1}^\bullet}} 
\newcommand{\NN}{\mathcal N} 
\newcommand{\goto}[1]{\mathrel{\raisebox{-2pt}{$\xrightarrow{#1}$}}} 
\newcommand{\Var}{\mathcal Var} 
\renewcommand{\empty}{{\gg}}

\newcommand{\bool}{\mathtt{bool}}
\newcommand{\integer}{\mathtt{int}}
\newcommand{\str}{\mathtt{string}}
\newcommand{\rational}{\mathtt{rat}}
\newcommand{\dom}{\textsc{dom}}
\newcommand{\E}{\mathcal{E}}		
\newcommand{\M}{\mathcal{M}}		
\newcommand{\PP}{\mathcal{P}}		
\newcommand{\restr}[2]{\left.#1\right|_{#2}} 
\newcommand{\cocomot}{CoCoMoT\xspace}
\newcommand{\tool}{\texttt{cocomot}\xspace}
\newcommand{\atom}{\mathit{Atoms}}
\newcommand{\equivcc}{\sim_{\mathit{cc}}}
\newcommand{\LQA}{\ensuremath{\mathcal{LQA}}\xspace}
\newcommand{\EUF}{\ensuremath{\mathcal{EUF}}\xspace}
\newcommand{\LIA}{\ensuremath{\mathcal{LIA}}\xspace}
\newcommand{\IDL}{\ensuremath{\mathcal{IDL}}\xspace}

\newcommand{\constraint}{c}
\newcommand{\runsof}[1]{Runs(#1)}
\newcommand{\firingsof}[1]{\mathcal{F}(#1)}
\newcommand{\moves}{Moves_{\NN}}
\newcommand{\alignment}[1]{Align(\NN,{#1})}
\newcommand{\OPTalignment}[1]{Align^{opt}(\NN,{#1})}

\newcommand{\eventsplus}{\E^{\empty}}
\newcommand{\firingsplus}{\mathcal{F}^{\empty}}

\newcommand{\procrun}{\vec f}
\newcommand{\logtrace}{\vec e}

\newcommand{\syncmove}[4]{\tikz[baseline=-0.5ex]{\node[scale=.6, inner sep=0pt]{$%
\begin{array}{|@{\,}l@{\quad}l@{\,}|}%
\hline{#1}&{#2}\\\hline{#3}&{#4}\\\hline\end{array}$}}}
\newcommand{\logmove}[2]{\tikz[baseline=-0.5ex]{\node[scale=.6, inner sep=0pt]{$%
\begin{array}{|@{\,}l@{\quad}l@{\,}|}%
\hline{#1}&{#2}\\\hline\multicolumn{2}{|c|}{\empty}\\\hline\end{array}$}}}
\newcommand{\modelmove}[2]{\tikz[baseline=-0.5ex]{\node[scale=.6, inner sep=0pt]{$%
\begin{array}{|@{\,}l@{\quad}l@{\,}|}%
\hline\multicolumn{2}{|c|}{\empty}\\\hline{#1}&{#2}\\\hline\end{array}$}}}
\newcommand{\modelmovenw}[1]{\tikz[baseline=-0.5ex]{\node[scale=.6, inner sep=0pt]{$%
\begin{array}{|@{\,}l@{\quad}l@{\,}|}%
\hline\multicolumn{2}{|c|}{\empty}\\\hline\multicolumn{2}{|c|}{#1}\\\hline\end{array}$}}}

\newcommand{\m}[1]{\mathsf{#1}}

\renewcommand{\todo}[2][]{}

\begin{document}
\maketitle

\begin{abstract}
Conformance checking is a key process mining task for comparing the expected behavior captured in a process model and the actual behavior recorded in a log. While this problem has been extensively studied for pure control-flow processes, conformance checking with multi-perspective processes is still at its infancy. In this paper, we attack this challenging problem by considering processes that combine the data and control-flow dimensions. In particular, we adopt data Petri nets (DPNs) as the underlying reference formalism, and show how solid, well-established automated reasoning techniques can be effectively employed for computing conformance metrics and data-aware alignments. We do so by introducing the CoCoMoT (Computing Conformance Modulo Theories) framework, with a fourfold contribution. First, we show how SAT-based encodings studied in the pure control-flow setting can be lifted to our data-aware case, using SMT as the underlying formal and algorithmic framework. Second, we introduce a novel preprocessing technique based on a notion of property-preserving clustering, to speed up the computation of conformance checking outputs. Third, we provide a proof-of-concept implementation that uses a state-of-the-art SMT solver and report on preliminary experiments. Finally, we discuss how CoCoMoT directly lends itself to a number of further tasks, like multi- and anti-alignments, log analysis by clustering, and model repair.
\end{abstract}

\section{Introduction}
\label{sec:intro}
In process mining, the task of conformance checking is crucial to match the 
expected behavior described by a process model against the actual action sequences
documented in a log \cite{CVSW18}.
While the problem has been thoroughly studied for pure control-flow processes such as classical Petri nets~\cite{Aalst11,CVSW18}, the situation changes for process models 
equipped with additional perspectives beyond the control-flow, such as for example the data perspective. In this inherently much more challenging setting, little research has been done on conformance checking, with few approaches focusing on declarative \cite{BuMS16} and procedural~\cite{MannhardtLRA16,Mannhardt18} multi-perspective process models with rather restrictive assumptions on the data dimension.

In this paper, we provide a new stepping stone in the line of research focused on conformance checking of multi-perspective procedural, Petri net-based process models. Specifically, we introduce a novel general framework, called \cocomot, to tackle conformance checking of data Petri nets (DPNs), an extensively studied formalism within BPM \cite{LeoniFM18,FeDM19} and process mining \cite{MDRV16,MannhardtLRA16,Mannhardt18}. The main feature of \cocomot is that, instead of providing ad-hoc algorithmic techniques for checking conformance, it provides an overarching approach based on the theory and practice of Satisfiability Modulo Theories (SMT).
By relying on an SMT backend, we employ well-established automated reasoning techniques that
can support data and operations from a variety of theories, restricting the data dimension as little as possible.

On top of this basis, we provide a fourfold contribution. First, 
we show that conformance checking of DPNs
can be reduced to satisfiability of an SMT formula over the theory of 
linear integer and rational arithmetic.
While our approach is inspired by the use of SAT solvers for a similar purpose~\cite{BoltenhagenCC21,ChatainCD17}, the use of SMT does not only allow us to support
data, but also capture unbounded nets.
Our \cocomot approach results in a conformance checking procedure running in NP,
which is optimal for the problem, in contrast to earlier approaches running in exponential time~\cite{MannhardtLRA16,Mannhardt18}.

Second, we show how to simplify and optimize conformance checking by introducing a preprocessing, trace clustering technique for DPNs that groups together traces that have the same minimal alignment cost. Clustering allows one to compute conformance metrics by just computing alignments of one representative per cluster, and to obtain alignments for other members of the same cluster as a simple adjustment of the alignment computed for the representative trace. 
Besides the
general notion of clustering, we then propose a concrete clustering strategy grounded in data abstraction for variable-to-constant
constraints, and show how this strategy leads to a significant speedup in our experiments.

Third, we report on a proof-of-concept implementation of \cocomot, discussing optimization techniques and showing the feasibility of the approach with an experimental evaluation on three different benchmarks.

Finally, we discuss how our approach, due to its modularity, directly lends itself to a number
  of further   process analysis tasks such as computing \emph{multi-} and \emph{anti-alignments}, using \cocomot as as a \emph{log clustering} method in the spirit of earlier work for Petri nets without data~\cite{ChatainCD17,BoltenhagenCC19b}, doing \emph{model repair}, and handling more sophisticated data such as persistent, relational data.

The remainder of the paper is structured as follows.
In Sec. \ref{sec:preliminaries} we recall the relevant basics about data Petri nets and alignments.
This paves the way to present our SMT encoding in Sec. \ref{sec:cocomot}.
Our clustering technique that serves as a preprocessor for conformance checking is the topic of \ref{sec:clustering}.
In Sec.\ref{sec:implementation} we describe our prototype implementation and the conducted experiments.
Afterwards, we discuss perspectives and potential of our approach in Sec. \ref{sec:discussion}. 

\section{Preliminaries}
\label{sec:preliminaries}

In this section we provide the required preliminaries from the relevant literature. We first recall data Petri nets (DPNs) and their execution semantics, then delve into event logs and conformance checking alignments, and finally discuss the main machinery behind our approach for satisfiability modulo theories (SMT). 

\subsection{Data Petri Nets}\label{sec:dpn}
We use Data Petri nets (DPNs) for modelling multi-perspective processes, adopting a formalization as in \cite{Mannhardt18,MannhardtLRA16}.
 
We start by introducing sorts -- data types of variables manipulated by a process.
We fix a set of \emph{(process variable) sorts} $\Sigma = \{\texttt{bool}, \texttt{int}, \texttt{rat},\texttt{string}\}$ with associated domains 
of booleans $\Dom(\bool) = \mathbb B$,
integers $\Dom(\integer) = \mathbb Z$,
rationals $\Dom(\rational) = \mathbb Q$, and 
strings $\Dom(\str) = \mathbb S$.
A set of \emph{process variables} $V$ is
\emph{sorted} if there is a function $sort\colon V \to \Sigma$ assigning a sort to each variable $v\in V$. 
For a set of variables $V$, we consider two disjoint sets of annotated variables $V^r = \{v^r \mid v\in V\}$ and $V^w = \{v^w \mid v\in V\}$ to be respectively read and written by process activities, as explained below, and we assume $sort(v^r) = sort(v^w) = sort(v)$ for every $v\in V$. 
For a sort $\sigma\in \Sigma$, 
$V_{\sigma}$ denotes the subset of $V^r\cup V^w$ of annotated variables of sort $\sigma$. 
To manipulate sorted variables, we consider expressions $\constraint$ with the following grammar: 
\begin{align*}
 c &= V_{\bool} \mid \mathbb B \mid n \geq n \mid r \geq r \mid r > r \mid s = s \mid b \wedge b \mid \neg b &
 s &= V_{\str} \mid \mathbb S \\
 n &= V_{\integer} \mid \mathbb Z \mid n + n \mid - n &
 r &= V_{\rational} \mid \mathbb Q \mid r + r \mid - r
\end{align*}

Standard equivalences apply, hence disjunction (i.e., $\lor$) and comparisons $\neq$, $<$, $\leq$ can be used as well ($\texttt{bool}$ and $\texttt{string}$ only support (in)equality).
These expressions form the basis to capture conditions on the values of variables that are read and written during the execution of activities in the process. For this reason, we call them \emph{constraints}. 
Intuitively, a constraint $(v_1^r > v_2^r)$ dictates that the current value of variable $v_1$ is greater than the current value of $v_2$. Similarly, $(v_1^w > v_2^r + 1) \land (v_1^w < v_3^r)$ requires that the new value given to $v_1$ (i.e., assigned to $v_1$ as a result of the execution of the activity to which this constraint is attached) is greater than the current value of $v_2$ plus $1$, and smaller than $v_3$. More in general, given a constraint $c$ as above, we refer to the annotated variables in $V^r$ and $V^w$ that appear in $\constraint$ as the \emph{read} and \emph{written variables}, respectively. 
The set of read and written variables that appear in a constraint $c$ is denoted by $\Var(\constraint)$, hence $\Var(\constraint)\subseteq V^w \cup V^r$. We denote the set of all constraints by $\mathcal C(V)$.

%


\begin{definition}[DPN]
A \emph{Petri net with data} (DPN) is given by a tuple
$\NN = (P, T, F, \ell, A, V,guard)$, where
\begin{inparaenum}[\it (1)]
\item $(P, T, F, \ell)$ is a Petri net with two non-empty disjoint sets of places $P$ and transitions $T$, 
a flow relation $F:(P \times T)\cup(T \times P)\rightarrow\mathbb{N}$ and a labeling injective function $\ell:T\to A\cup \set{\tau}$, where $A$ is a finite set of activity labels and $\tau$ is a special symbol denoting silent transitions;
\item $V$ is a sorted set of process variables; and
\item $guard\colon T \to \mathcal C(V)$ is a guard assignment. 
\end{inparaenum} 
\end{definition}

As customary, given $x\in P \cup T$, we use $\pre{x}:=\set{y\mid F(y,x)>0}$ to denote the \emph{preset} of $x$ and $\post{x}:=\set{y\mid F(x,y)>0}$ to denote the \emph{postset} of $x$. 
In order to refer to the variables read and written by a transition $t$,
we use the notations $read(t) = \{v \mid v^r \in \Var(guard(t)) \}$ and
$write(t) = \{v \mid v^w \in \Var(guard(t)) \}$. Finally, $G_\NN$ is the set of all the guards appearing in $\NN$.

To assign values to variables, we use variable assignments. A \emph{state variable assignment} is a total function $\alpha$ that assigns a value to each variable in $V$, namely $\alpha(v)\in \Dom(sort(v))$ for all $v\in V$. These assignments are used to specify the current value of all variables. 
Similarly, a \emph{transition variable assignment} is a partial function $\beta$ that assigns a value to annotated variables, namely $\beta(x)\in \Dom(sort(x))$, with $x\in V^r\cup V^w$. These are used to specify how variables change as the result of activity executions (cf. Def.~\ref{def:tfiring}). 

A \emph{state} in a DPN $\NN$ is a pair $(M,\alpha)$ constituted by a marking $M:P\rightarrow\mathbb{N}$ for the underlying petri net $(P, T, F, \ell)$, plus a state variable assignment. A state thus simultaneously accounts for the control flow progress and for the current values of all variables in $V$, as specified by $\alpha$.

We now define when a Petri net transition may fire from a given state. 

\begin{definition}[Transition firing]
\label{def:tfiring}
A transition $t\in T$ is \emph{enabled} in state $(M, \alpha)$ if a transition variable assignment $\beta$ exists such that:
\begin{compactenum}[\it (i)]
\item  $\beta(v^r) = \alpha(v)$ for every $v\in read(t)$, i.e., $\beta$ is as $\alpha$ for read variables;
\item  $\beta \models guard(t)$, i.e., $\beta$ satisfies the guard; and 
\item  $M(p) > F(p,t)$ for every $p \in \pre{t}$.
\end{compactenum}
An enabled transition may \emph{fire}, producing a new state $(M',\alpha')$, s.t. $M'(p) = M(p) - F(p,t) + F(t,p)$ for every $p\in P$, and $\alpha'(v) = \beta(v^w)$ for every $v\in write(t)$, and $\alpha'(v) = \alpha(v)$ for every $v\not\in write(t)$. 
A pair $(t,\beta)$ as above is called (valid) \emph{transition firing}, and we denote its firing by $(M, \alpha) \goto{(t, \beta)} (M', \alpha')$.
\end{definition}

\noindent
Given $\NN$, we fix one state $(M_I,\alpha_0)$ as \emph{initial}, where $M_I$ is the initial marking of the underlying Petri net $(P, T, F, \ell)$ and $\alpha_0$ specifies the initial value of all variables in $V$.  Similarly, we denote the final marking as $M_F$, and call \emph{final} any state of $\NN$ of the form $(M_F,\alpha_F)$ for some $\alpha_F$. 

We say that $(M',\alpha')$ is \emph{reachable} in a DPN iff there exists a sequence of transition firings
$\procrun =(t_1, \beta_1), \dots, (t_n, \beta_n)$, s.t. $(M_I,\alpha_0)\goto{(t_1, \beta_1)} \ldots\goto{(t_n, \beta_n)}(M',\alpha')$,  denoted as $(M_I, \alpha_0) \goto{\procrun} (M_n, \alpha_n)$. 
Moreover, $\procrun$ is called a (valid) \emph{process run} of $\NN$ if $(M_I,\alpha_0)\goto{\procrun} (M_F,\alpha_F)$ for some $\alpha_F$. 
%
Similar to~\cite{MannhardtLRA16}, we restrict to \emph{relaxed data sound} DPNs, that is, 
where at least one final state is reachable.  

We denote the set of valid transition firings of a DPN $\NN$ as $\firingsof{\NN}$, and the set of process runs as $\runsof{\NN}$.  

\begin{example}
\label{exa:1}
Consider the following DPN $\NN$:
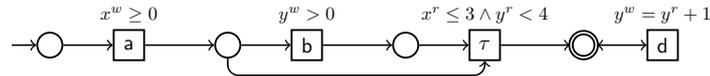
\begin{center}
\begin{tikzpicture}[node distance=13mm]
\tikzstyle{place}=[draw, circle, line width=.7pt]
\tikzstyle{trans}=[draw, rectangle, line width=.7pt, scale=.8, minimum height=5mm, minimum width=5mm]
\tikzstyle{goto}=[->, line width=.6pt]
\tikzstyle{tlabel}=[yshift=4mm, scale=.7]
\node[place] (p0) {};
\node[trans, right of=p0] (a) {\textsf{a}};
\node[tlabel] at (a) {$x^w \geq 0$};
\node[place, right of=a] (p1) {};
\node[trans, right of=p1] (b) {\textsf{b}};
\node[tlabel] at (b) {$y^w > 0$};
\node[place, right of=b] (p2) {};
\node[trans, right of=p2] (c) {$\tau$};
\node[tlabel] at (c) {$x^r \leq 3 \wedge y^r < 4$};
\node[place, right of=c, double] (p3) {};
\node[trans, right of=p3] (d) {\textsf{d}};
\node[tlabel] at (d) {$y^w = y^r + 1$};
\draw[goto] ($(p0) + (-.5,0)$) -- (p0);
\draw[goto] (p0) -- (a);
\draw[goto, rounded corners] (p1) -- ($(p1) - (0,.4)$) -- ($(c) - (0,.4)$) -- (c);
\draw[goto] (a) -- (p1);
\draw[goto] (p1) -- (b);
\draw[goto] (b) -- (p2);
\draw[goto] (p2) -- (c);
\draw[goto] (c) -- (p3);
\draw[goto, <->] (p3) -- (d);
\end{tikzpicture}
\end{center}
The set $\runsof{\NN}$ contains, e.g.,
$\langle (\mathsf{a}, \{x^w\mapsto 2\}), (\mathsf{b}, \{y^w\mapsto 1\}), (\tau, \varnothing)\rangle$ and
$\langle (\mathsf{a}, \{x^w\mapsto 1\}), (\tau, \varnothing), (\mathsf{d}, \{y^w\mapsto 1\})\rangle$, for $\alpha_0 = \{x\mapsto 0, y\mapsto 0\}$.
\end{example}

\subsection{Event Logs and Alignments}
\label{sec:alignments}

Given an arbitrary set  $A$ of activity labels, an \emph{event} is a pair $(b,\alpha)$, where $b\in A$ and $\alpha$ is a so-called \emph{event variable assignment} (which, differently from state variable assignments, can be a partial function). 

\begin{definition}[Log trace, event log]
Given a set $\E$ of events, a \emph{log trace} $\logtrace\in\E^*$ is a sequence of events in $\E$ and an \emph{event log} $L\in\M(\E^*)$ is a multiset of log traces from $\E$, where  $\M(\E^*)$ denotes the set of multisets over $\E^*$.
\end{definition}


We focus on a conformance checking procedure that aims at constructing an \emph{alignment} of a given log trace $\logtrace$ w.r.t. the process model (i.e., the DPN $\NN$), by matching events in the log trace against transitions firings in the process runs of $\NN$. 
However, when constructing an alignment, not every event can always be put in correspondence with a transition firing, and vice versa. 
Therefore, we introduce a special ``skip" symbol $\empty$ and the extended set of events $\eventsplus = \E\cup \set{\empty}$ and, given $\NN$, the extended set of transition firings $\firingsplus=\firingsof{\NN}\cup\set{\empty}$. 

%
%

Given a DPN $\NN$ and a set  $\E$ of events as above, a pair $(e,f)\in \eventsplus \times \firingsplus \setminus \{(\empty,\empty)\}$ is called \emph{move}.%
\footnote{In contrast to~\cite{MannhardtLRA16}, we do not here 
distinguish between synchronous moves with correct and
incorrect write operations, but defer this differentiation to the cost function.} 
A move $(e,f)$ is called: 
\begin{inparaenum}[(i)]
\item \emph{log move} if $e \in \E$ and $f = \empty$; 
\item \emph{model move} if $e = \empty$ and $f \in \firingsof{\NN}$; 
\item \emph{synchronous move} if $(e,f) \in \E \times \firingsof{\NN}$. 
\end{inparaenum}
Let $\moves$ be the set of all such moves. We now show how moves can be used to define an alignment of log trace.

For a sequence of moves 
$\gamma = (e_1,f_1), \dots, (e_n,f_n)$,
the \emph{log projection} $\restr{\gamma}{L}$ of $\gamma$ is the maximal subsequence $\seq[i]{e'}$ of $\seq[n]e$ such that $\seq[i]{e'}\in \E^*$, that is, it contains no $\empty$ symbols.
Similarly, the \emph{model projection} $\restr{\gamma}{M}$ of $\gamma$ is the maximal subsequence $\seq[j]{f'}$ of $\seq[n]f$ such that $\seq[j]{f'}\in \firingsof{\NN}^*$.

\begin{definition}[Alignment]
Given $\NN$, a sequence of legal moves $\gamma$
is an \emph{alignment} of a log trace $\logtrace$ if $\restr{\gamma}{L} = \logtrace$, and 
it is \emph{complete} if $\restr{\gamma}{M}\in\runsof{\NN}$.
\end{definition}

\begin{example}
\label{exa:2}
The trace $\logtrace = \langle (\mathsf{a}, \{x^w\mapsto 2\}), (\mathsf{b}, \{y^w\mapsto 2\})\rangle$ has the following alignments in the DPN from Ex. \ref{exa:1}:
\[
\gamma_1=
\syncmove{\m a}{x=2}{\m a}{x^w=2}
\syncmove{\m b}{y=1}{\m b}{y^w=1}
\modelmovenw{\tau}
\qquad
\gamma_2=
\syncmove{\m a}{x=2}{\m a}{x^w=3}
\modelmovenw{\tau}
\logmove{\m b}{y=1}
\qquad
\gamma_3=
\logmove{\m a}{x=2}
\logmove{\m b}{y=1}
\modelmove{\m a}{x^w=3}
\modelmovenw{\tau}
\]
\end{example}

We denote by $\alignment{\logtrace}$ the set of complete alignments for a log trace $\logtrace$ w.r.t. $\NN$.
A \emph{cost function} is a mapping $\kappa\colon \moves \to \mathbb R^+$ that assigns a cost to every move. It is naturally extended to alignments as follows. 

\begin{definition}[Cost]
Given $\NN$, $\logtrace$ and $\gamma \in \alignment{\logtrace}$ as before, the \emph{cost} of $\gamma$ is obtained by summing up the costs
of its moves, that is,
$\kappa(\gamma) = \sum_{i=1}^n \kappa(e_i,f_i)$.
Moreover, $\gamma$ is
\emph{optimal for $\logtrace$} if $\kappa(\gamma)$ is minimal among all complete alignments for $\logtrace$, namely there is no $\gamma'\in \alignment{\logtrace}$ with $\kappa(\gamma')<\kappa(\gamma)$. 
\end{definition}

\noindent
We denote the cost of an optimal alignment for $\logtrace$ with respect to 
$\NN$ by $\kappa_\NN^{opt}(\logtrace)$. Given $\NN$, the set of optimal alignments for $\logtrace$ is denoted by $\OPTalignment{\logtrace}$.

\subsection{Satisfiability Modulo Theories (SMT)}
\label{sec:smt}
We assume the usual syntactic (e.g., signature, variable, term, atom, literal, and formula) and semantic (e.g., structure, truth, satisfiability, and validity) notions of first-order
logic. The equality symbol $=$ is always included in all signatures.
 An expression is a term, an atom, a literal, or
a formula. Let $\underline{x}$ be a finite tuple of variables and $\Sigma$ a signature; a $\Sigma(\underline{x})$-expression is an
expression built out of the symbols in $\Sigma$ where only (some of) the variables in $\underline{x}$ may occur free
(we write $E(\underline{x})$ to emphasize that E is a $\Sigma(\underline{x})$-expression).

According to the current practice in the SMT literature \cite{smt-lib}, a theory $\mathcal{T}$ is a pair $(\Sigma, Z)$, 
where $\Sigma$ is a signature and $Z$ is a class of $\Sigma$-structures; the structures in $Z$ are the models
of $T$. We assume $\mathcal{T} = (\Sigma, Z)$. A $\Sigma$-formula 
$\phi$ is $T$-satisfiable if there exists a $\Sigma$-structure
$\mathcal{M}$ in $Z$ such that $\phi$ is true in $\mathcal{M}$ under a suitable assignment $\mathtt{a}$ to the free variables of $\phi$ (in
symbols, $(\mathcal{M}, \mathtt a) \models \phi$); it is $\mathcal{T}$-valid (in symbols, $T\vdash \phi$) if its negation is $\mathcal{T}$-unsatisfiable. Two
formulae $\phi_1$ and $\phi_2$ are $\mathcal{T}$-equivalent if $\phi_1\leftrightarrow\phi_2$ is $\mathcal{T}$-valid. The problem of (quantifier-free) \emph{satisfiability
modulo the theory $\mathcal{T}$} ($SMT(\mathcal{T})$) amounts to establishing the $\mathcal{T}$-satisfiability of
quantifier-free $\Sigma$-formulae. 

Intuitively, the \textit{Satisfiability Modulo Theories} (SMT) problem is a decision problem for the satisfiability of quantifier-free first-order formulae that extends the problem of propositional (boolean) satisfiability (SAT) by taking into account (the combination of) background first-order theories (e.g., arithmetics, bit-vectors, arrays, uninterpreted functions). There exists a plethora of solvers, called \emph{SMT solvers}, able to solve the SMT problem: they extend SAT-solvers with specific decision procedures customized for the specific theories involved. SMT solvers are useful both for computer-aided verification, to prove the correctness of software programs against some property of interest, and for synthesis, to generate candidate program fragments. Examples of well-studied SMT theories  are 
the theory of uninterpreted functions $\EUF$, the theory of bitvectors $\mathcal{B}\mathcal{V}$ and the theory of arrays $\mathcal{A}\mathcal{X}$. All these theories are usually employed in applications to program verification. SMT solvers also support different types of arithmetics for which specific decision procedures are available, like difference logic $\IDL$ (whose atoms are of the form $x-y\leq c$ for some integer constant $c$), or linear arithmetics ($\LIA$ for integers and $\LQA$ for rationals). In this paper we will focus on $\EUF$, $\LIA$ and $\LQA$, since our constraint language can be expressed having as background the combination of such theories.

Another important problem studied in the SMT literature is the one of Optimization Modulo Theories (OMT). OMT is an extension of SMT, whose goal is to
find models that make a given objective optimum through a combination of SMT and
optimization procedures. In this paper we will consider a sub-case of OMT, that is called \textsc{MaxSMT}, where the task is to maximize/minimize a given function. 

SMT-LIB  \cite{smt-lib} is an international initiative with the aims of providing  an extensive on-line library of benchmarks and of
promoting the adoption of common languages and interfaces for SMT solvers. For the purpose of this paper, we make use of the Yices SMT solvers \cite{yices,Dutertre14} (version 2.6.2) and Z3 \cite{deMouraB08}.


\section{Conformance Checking via SMT}
\label{sec:cocomot}

In this section we illustrate our approach. 
We first describe in Section \ref{ssec:distance} a generic distance measure to be used as cost function. Then, in Section \ref{ssec:encoding} detail our encoding of the problem of finding optimal alignments in SMT. 
Notably, this technique works also for nets with arc multiplicities and unbounded nets, beyond the bounded case considered in \cite{BoltenhagenCC21}\todo{CHECK!}. 
Finally, in Section~\ref{ssec:complexity} we analyze the computational complexity of our approach. 

\subsection{Distance-based Cost Function}
\label{ssec:distance}

We present here a function used to measure the distance between a log trace and a process run.
The recursive definition has the same structure as that of the standard edit distance, which allows us to adopt a similar encoding as used in the literature~\cite{BoltenhagenCC19}.
However, it generalizes both the standard edit distance and distance functions previously used for multi-perspective conformance checking~\cite{MannhardtLRA16,Mannhardt18}, and admits also other measures that are specific to the model and the SMT theory used.
Our measure is parameterized by three functions:
\[
P_L\colon \E \to \mathbb N \qquad
P_M\colon \firingsof{\NN} \to \mathbb N \qquad 
P_=\colon \E \times \firingsof{\NN} \to \mathbb N
\]
respectively called the \emph{log move penalty}, \emph{model move penalty}, and
\emph{synchronous move penalty} functions (cf. Section~\ref{sec:alignments}). 
We use these functions to assign penalties to log moves, model moves, or synchronous moves, respectively. 
In what follows, we denote prefixes of length $j$ of a log trace $\logtrace \in \E^*$ of length $m$ as $\logtrace|_j$, provided $0\leq j\leq m$, and analogously for a process run $\procrun\in\runsof{\NN}$ (recall that these are sequences of transition firings in $\firingsof{\NN}$).

\begin{definition}[Edit distance]
\label{def:distance}
Given a DPN $\NN$, let
$\logtrace = \seq[m]e$ be a log trace and
$\procrun = \seq f$ a process run.
For all $i$ and $j$, $0\,{\leq}\,i\,{\leq}\,m$ and $0\leq j \leq n$,
the \emph{edit distance} 
$\dist(\logtrace|_{i}, \procrun|_{j})$
is recursively defined as follows:
\begin{align*}
\dist(\epsilon, \epsilon) &= 0 \\
\dist(\logtrace|_{i+1}, \epsilon) &= P_L(e_{i+1}) + \dist(\logtrace|_{i}, \epsilon)
\\
\dist(\epsilon, \procrun|_{j+1}) &=  P_M(f_{j+1}) + \dist( \epsilon,\procrun|_{j}) 
\\
\dist(\logtrace|_{i+1}, \procrun|_{j+1}) &=
\min\left\{
\begin{array}{l}
\dist(\logtrace|_{i}, \procrun|_{j}) + P_=(e_{i+1}, f_{j+1}) \\
P_L(e_{i+1}) + \dist(\logtrace|_{i}, \procrun|_{j+1}) \\
P_M(f_{j+1}) + \dist(\logtrace|_{i+1}, \procrun|_{j})          
\end{array}\right.
\end{align*}
\end{definition}

\noindent
Def.~\ref{def:distance} can be used to define a cost function by setting 
$\kappa(\gamma) = \delta(\restr\gamma L, \restr\gamma M)$, for any alignment $\gamma$.
In the sequel, we call such a cost function \emph{distance-based}.
Moreover, it is known that for any trace $\logtrace$ and process run $\procrun$ with $|\logtrace|=m$ and $|\procrun| = n$, given the $(n+1) \times (m+1)$-matrix $D$ such that
$D_{ij} = \delta(\logtrace|_{i}, \procrun|_{j})$, one can reconstruct an alignment
of $\logtrace$ and $\procrun$ that is optimal with respect to $\kappa$ \cite{NeedlemanW70,BoltenhagenCC21}.

\begin{remark}
\label{rem:distances}
By fixing the parameters $P_=$, $P_L$, and $P_M$ of Def.~\ref{def:distance}, 
one obtains concrete, known distance-based cost functions, such as the following:

\noindent
\textbf{Standard cost function.} 
Def.~\ref{def:distance} can be instantiated to the measure in \cite[Ex. 2]{MannhardtLRA16}, \cite[Def. 4.5]{Mannhardt18}.
To that end, we set
$P_L(b,\alpha) = 1$;
$P_M(t,\beta) = 0$ if $t$ is silent (i.e., $\ell(t) = \tau$) and 
$P_M(t,\beta) = |write(t)| + 1$ otherwise; and
$P_=((b, \alpha), (t,\beta)) = |\{v\in \dom(\alpha) \mid \alpha(v) \neq \beta(v^w)\}|$ if $b = \ell(t)$ and
$P_=((b, \alpha), (t,\beta)) = \infty$ otherwise.

\noindent
\textbf{Levenshtein distance.} 
The standard edit distance is obtained with 
$P_L(b, \alpha) =  P_M(t, \beta) = 1$, and
$P_=((b, \alpha), (t,\beta)) =  0$ if $b=\ell(t)$ and $P_=((b, \alpha), (t,\beta)) =  \infty$ otherwise.
Note that this measure ignores transition variable assignments $\beta$.
\end{remark}

For instance, for the alignments $\gamma_1$, $\gamma_2$, and $\gamma_3$ from Ex. \ref{exa:2}, the standard cost function yields $\kappa(\gamma_1) = 0$; $\kappa(\gamma_2) = 2$ (because we get penalty $1$ for a synchronous move with incorrect write operation, no penalty for the invisible model move, and penalty $1$ for the log move); and $\kappa(\gamma_3) = 4$ (because we get penalty $1$ for each of the log moves, penalty $2$ for a visible model move that writes one variable, and no penalty for the invisible model move).

\subsection{Encoding}
\label{ssec:encoding}

Our approach relies on the fact that the an
optimal alignment for a given log trace is upper-bounded in length.
To this end, we use the following observation.\todo{Marco: we have to say something about proofs. Like at the end of the intro "For sake of space we do not porovide proofs here, they can be found in a technical report \cite{blah}".}

\begin{remark}
\label{rem:bound}
Given a DPN $\NN$ and a log trace $\logtrace = \seq[m]e$,
let $\procrun = \seq[n]f$ be a valid process run such that 
$\sum_{j=1}^n P_M(f_j)$ is minimal.
Then an optimal alignment $\gamma$ for $\logtrace$ and $\NN$ satisfies
$\kappa(\gamma) \leq \kappa(\gamma_{max})$, and hence $|\gamma| \leq |\gamma_{max}|$, where $\gamma_{max}$
is the alignment
$
(e_1, \empty), \dots, (e_m, \empty), (\empty, f_1), \dots (\empty, f_n)
$.
\end{remark}

\noindent
Given a log trace $\logtrace = \seq[m]e$ and a DPN $\NN$ with initial marking $M_I$, initial state variable assignment $\alpha_0$, final marking $M_F$, we want to construct an optimal alignment $\gamma \in \OPTalignment{\logtrace}$.
To that end, we assume throughout this section that the number of non-empty model steps in $\gamma$ is bounded by $n$  (cf. Rem. \ref{rem:bound}).
Our approach comprises the following four steps:
(1) represent the alignment symbolically by a set of SMT variables,
(2) set up constraints $\Phi$ that symbolically express optimality of this alignment,
(3) solve the constraints $\Phi$ to obtain a satisfying assignment $\nu$, and
(4) decode an optimal alignment $\gamma$ from $\nu$.
We next elaborate these steps in detail.
\smallskip

\noindent
\textbf{(1) Alignment representation.}
We use the following SMT variables:
\begin{itemize}
\item[(a)]
transition step variables $S_i$ for $1\leq i \leq n$
of type integer; if $T = \{t_1, \dots, t_{|T|}\}$ then it is ensured
that $1\leq S_i \leq |T|$, with the semantics that
$S_i$ is assigned $j$ iff the $i$-th
transition in the process run is $t_j$;
\item[(b)]
marking variables $M_{i,p}$ of type integer for all $i$, $p$ with
$0\leq i \leq n$ and $p\in P$, where $M_{i,p}$ is assigned $k$ iff
there are $k$ tokens in place $p$ at instant $i$;
\item[(c)]
data variables $X_{i,v}$ for all $v\in V$ and $i$, $0\leq i \leq n$;
the type of these variables depends on $v$, 
with the semantics that $X_{i,v}$ is assigned $r$ iff the value of $v$ at 
instant $i$ is $r$;
we also write $X_{i}$ for $(X_{i,v_1}, \dots, X_{i,v_k})$;
\item[(d)]
distance variables $\delta_{i,j}$ of type integer
for $0\leq i \leq m$ and $0\leq j \leq n$, where
$\delta_{i,j}=d$ if $d$ is the cost of
the prefix $\logtrace|_i$ of the log trace $\logtrace$, and the prefix $\procrun|_j$ of the (yet to be determined)
process run $\procrun$, i.e., $d = \delta(\logtrace|_i, \procrun_j)$ by Def. \ref{def:distance}.
\end{itemize}
\noindent
Note that variables (a)--(c) comprise all information required to capture a process run with $n$ steps, which will make up the model projection
of the alignment $\gamma$, while the distance variables (d) will be used to
encode the alignment.
\smallskip

\noindent
\textbf{(2) Encoding.}
To ensure that the values of variables correspond to a valid run,
we assert the following constraints:
\begin{compactitem}[$\bullet$]
\item 
The initial marking $M_I$ and the initial assignment $\alpha_0$ are respected:
\begin{align}
\label{eq:init}
\tag{$\varphi_{\mathit{init}}$}
\textstyle\bigwedge_{p\in P} M_{0,p} = M_I(p) \wedge
\bigwedge_{v\in V} X_{0,v} = \alpha_0(v)
\end{align}
\item 
The final marking $M_F$ is respected:
\begin{align}
\label{eq:final marking}
\tag{$\varphi_{\mathit{final}}$}
\textstyle\bigwedge_{p\in P} M_{n,p} = M_F(p)
\end{align}
\item
Transitions correspond to transition firings in the DPN:
\begin{align}
\label{eq:transition range}
\tag{$\varphi_{\mathit{trans}}$}
\textstyle\bigwedge_{1\leq i \leq n} 1 \leq S_i \leq |T|
\end{align}
In contrast to \cite{BoltenhagenCC19}, no constraints are needed to
express that at every instant exactly one transition occurs, since the value of $S_i$ is unique.
\item Transitions are enabled when they fire:
\begin{align}
\label{eq:enabled}
\tag{$\varphi_{\mathit{enabled}}$}
\textstyle\bigwedge_{1\leq i \leq n} \bigwedge_{1\leq j \leq |T|}
{(S_i\,{=}\,j)} \to 
\bigwedge_{p\,\in\,\pre{t_j}} M_{i-1,p} \geq |\pre{t_j}|_p 
\end{align}
where $|\pre{t_j}|_p$ denotes the multiplicity of $p$ in the multiset $\pre{t_j}$.
\item We encode the token game:
\begin{align}
\label{eq:token game}
\tag{$\varphi_{\mathit{mark}}$}
\bigwedge_{1\leq i \leq n} \bigwedge_{1\leq j \leq |T|}
{(S_i\,{=}\,j)} \to 
\bigwedge_{p\,\in\,P} M_{i,p} - M_{i-1,p} =  |\post{t_j}|_p -  |\pre{t_j}|_p 
\end{align}
where $|\post{t_j}|_p$ is the multiplicity of $p$ in the multiset $\post{t_j}$.
\item The transitions satisfy the constraints on data:
\begin{align}
\label{eq:data}
\tag{$\varphi_{\mathit{data}}$}
\bigwedge_{1\leq i < n} \bigwedge_{1\leq j \leq |T|}
{(S_i\,{=}\,j)} \to 
guard(t_j)\chi \wedge
\bigwedge_{v\not \in write(t_j)} X_{i-1,v} = X_{i,v}
\end{align}
where the substitution $\chi$ uniformly replaces $V^r$ by $X_{i-1}$ and $V^w$ by $X_i$.
\item
The encoding of the data edit distance depends on the penalty functions $P_=$, $P_M$, and $P_L$. We illustrate here the formulae obtained for the standard cost function in Rem. \ref{rem:distances}.\todo{Marco: What is (1)? The standard cost? Sarah: yes, added standard}
Given a log trace 
$\logtrace = (b_1,\alpha_1), \dots, (b_m,\alpha_m)$, let
the expressions $[P_L]$, $[P_M]_{j}$, and $[P_=]_{i,j}$
be defined as follows, for all $i$ and $j$:
\begin{align*}
[P_L] &= 1 \\
[P_M]_{j} &=
ite(S_j = 1, c_w(t_1), \dots ite(S_j = {|T|-1}, c_w(t_{|T|-1}), c_w(t_{|T|}))\dots)\\
[P_=]_{i,j} &= 
ite(S_j = b_i, \sum_{v\in write(b_i)} ite(\alpha_i(v) = X_{i,v}, 0, 1), \infty)
\end{align*}
where the \emph{write cost} $c_w(t)$ of transition $t\in T$ is 0 if $\ell(t) = \tau$,
or $|write(t)| + 1$ otherwise, and $ite$ is the if-then-else operator.
It is then straightforward to encode the data edit distance 
by combining all equations in Def. \ref{def:distance}:
\begin{equation*}
\begin{array}{rl@{\qquad}rl@{\qquad}rl@{\qquad\quad}r}
\dist_{0,0} &= 0 &
\dist_{{i+1},0} &= [P_L] + \dist_{i,0} &
\dist_{0,{j+1}} &= [P_M]_{j+1} + \dist_{0,j} 
&\hfill(\varphi_\delta)\\[1ex] 
\dist_{i+1,j+1} &\multicolumn{5}{l}{=
\min (
[P_=]_{i+1, j+1} + \dist_{i,j},\ 
[P_L] + \dist_{i,j+1},\ 
[P_M]_{j+1} + \dist_{i+1,j})}
\end{array}
\end{equation*}
\end{compactitem}

\noindent
\textbf{(3) Solving.}
We use an SMT solver to obtain a satisfying assignment $\nu$ for the 
following constrained optimization problem:
\begin{align}
\label{eq:constraints}
\varphi_{\mathit{init}} \wedge
\varphi_{\mathit{final}} \wedge
\varphi_{\mathit{trans}} \wedge
\varphi_{\mathit{enabled}} \wedge
\varphi_{\mathit{mark}} \wedge
\varphi_{\mathit{data}} \wedge
\varphi_{\delta}
\text{\quad minimizing\quad }\delta_{m,n}
\tag{$\Phi$}
\end{align}

\noindent
\textbf{(4) Decoding.}
We obtain a valid process run $\procrun =  f_1, \dots, f_n $
by decoding with respect to $\nu$
the variable sets 
$S_i$ (to get the transitions taken), $M_{i,p}$ (to get the markings), and $X_{i,v}$ (to get the state variable assignments) for every instant $i$, 
as described in Step (1).
Moreover, we use the known correspondence between edit distance and alignments~\cite{NeedlemanW70}
to reconstruct an alignment $\gamma = \gamma_{m,n}$ of $\logtrace$ and $\procrun$.
To that end, consider the (partial) alignments $\gamma_{i,j}$ recursively defined as follows:
\begin{align*}
\gamma_{0,0} &=\epsilon \qquad
\gamma_{i+1,0}= \gamma_{i,0} \cdot (e_{i+1}, \empty) \qquad
\gamma_{0,j+1}= \gamma_{0,j} \cdot (\empty, f_{j+1}) \\
\gamma_{i+1,j+1} &= 
\begin{cases}
\gamma_{i,j+1} \cdot (e_{i+1}, \empty) &
 \text{ if }\nu(\delta_{i+1,j+1}) = \nu([P_L] + \dist_{i,j+1}) \\
\gamma_{i+1,j} \cdot (\empty, f_{j+1}) &
 \text{ if otherwise }\nu(\delta_{i+1,j+1}) = \nu([P_M]_{j+1} + \dist_{i+1,j}) \\
\gamma_{i,j} \cdot (e_{i+1}, f_{j+1}) &
 \text{ otherwise}
\end{cases}
\end{align*}

\noindent
To obtain an optimal alignment, we use the following result:

\begin{theorem}
Let $\NN$ be a DPN, $\logtrace$ a log trace and $\nu$ a solution to \eqref{eq:constraints}.
Then $\gamma_{m,n}$ is an optimal alignment for $\logtrace$, i.e., $\gamma_{m,n}\in \OPTalignment{\logtrace}$.
\end{theorem}

\subsection{Complexity}
\label{ssec:complexity}

In this section we briefly comment on the computational complexity
of our approach and the (decision problem version of the) 
optimal alignment problem.
To that end, let a cost function $\kappa$ be 
\emph{well-behaved} if it is distance-based and its parameter functions
$P_=$, $P_M$, and $P_L$ are effectively computable and
can be defined by linear arithmetic expressions and case distinctions.
For $c\in \mathbb N$ and a well-behaved cost function $\kappa$, let $\textsc{Align}_{c}$ be the problem that, 
given a relaxed data-sound DPN and a log trace,
checks whether an alignment of cost $c$ with respect to $\kappa$ exists.
For any given DPN $\NN$, log trace $\logtrace$ and cost $c$, the encoding
presented in Sec.~\ref{ssec:encoding} is used to construct an SMT problem
over linear integer/rational arithmetic that is satisfiable if and only if an alignment of cost $c$ exists. 
The size of such an encoding is polynomial in the size of the DPN and the length of the log trace.
Thus, since satisfiability of the relevant class of SMT problems is in
NP~\cite{BradleyManna2007}, our approach to decide $\textsc{Align}_{c}$ is in NP.
In contrast, the approach presented in~\cite{MannhardtLRA16,Mannhardt18} 
is 
exponential in the length of the log trace.
Moreover, $\textsc{Align}_c$ is NP-hard since it is easy to reduce satisfiability of a boolean formula (SAT) to $\textsc{Align}_0$. Hence, all in all $\textsc{Align}_c$ is NP-complete.
\todo[inline]{use remainder of subsection only in long version}\todo{MArco: what about stating complexity as a theorem, so that the text above is just a proof sketch, and the text below can indeed be eliminated from the conference version?}
Given a boolean formula $\varphi$ with variables $V$, 
let $\NN_\varphi$ be the following DPN:
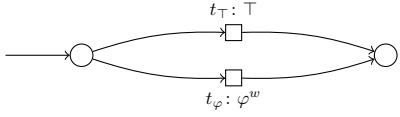
\begin{center}
\begin{tikzpicture}[node distance=3mm]
\tikzstyle{place}=[draw, circle, inner sep=3pt]
\tikzstyle{transition}=[draw, rectangle, inner sep=3pt]
\tikzstyle{tlabel}=[scale=.7]
\node[place] (p0) {};
\node[place] (p1) at (4,0) {};
\node[transition] (ttop) at (2,.3) {};
\node[above of=ttop, tlabel] {$t_\top \colon \top$};
\node[transition] (tphi) at (2,-.3) {};
\node[below of=tphi, tlabel] {$t_\varphi \colon \varphi^w$};
\draw[->] (p0) to[bend left=10] (ttop);
\draw[->] (p0) to[bend right=10] (tphi);
\draw[->] (ttop) to[bend left=10] (p1);
\draw[->] (tphi) to[bend right=10] (p1);
\draw[->] ($(p0) + (-1,0)$) -- (p0);
\end{tikzpicture}
\end{center}
where $\varphi^w$ is the formula obtained from $\varphi$ by replacing
all variables $v\in V$ by $v^w$.
The DPN $\NN_\varphi$ is relaxed data-sound due to the transition $t_\top$.
Let $\logtrace$ be the log trace consisting of the single event $(t_\varphi, \emptyset)$, and
$\kappa$ the standard edit distance (cf. Rem. \ref{rem:distances}).
Note that $\runsof{\NN_\varphi}$ contains at most two valid process runs:
we have $\procrun_0 = (t_\top, \varnothing) \in \runsof{\NN_\varphi}$ and $\kappa(\logtrace, \procrun_0) = \infty$.
If $\varphi$ is satisfiable by some assignment $\alpha$, we also have
$\procrun_1 = (t_\varphi, \alpha_w) \in \runsof{\NN_\varphi}$,
where $\alpha_w$ is the assignment such that $\alpha(v) = \alpha_w(v^w)$
for all $v\in V$, and $\kappa(\logtrace, \procrun_1) = 0$.
Thus, $\logtrace$ admits an alignment of cost $0$ if and only if $\varphi$
is satisfiable.

\section{Trace Clustering}
\label{sec:clustering}
\todo[inline]{add more citations on papers doing log clustering}

Clustering techniques are used to group together multiple portions of a process log in order to optimize their  analysis~\cite{BoltenhagenCC19b,ChatainCD17}.
In this section we provide means to simplify conformance checking of a log $L$ in a preprocessing phase, where the log $L$ is partitioned into groups with the same cost optimal alignment.

We express such partitioning by means of an equivalence relation $\equiv$ on the log traces in a log $L$, which thus identifies equivalence classes called \emph{clusters}.
%

\begin{definition}[Cost-based clustering]
Given a DPN $\NN$, a log $L$, and a cost function $\kappa$,
a \emph{cost-based clustering}
is an equivalence relation $\equiv_{\kappa_\NN^{opt}}$ over $L$,
where, for all traces $\tau,\tau'\in L$ s.t. $\tau\equiv_{\kappa_\NN^{opt}}\tau'$ we have that 
$\kappa_\NN^{opt}(\tau) = \kappa_\NN^{opt}(\tau')$.
\end{definition}

\noindent
%
We now introduce one specific equivalence relation that focuses on
DPN guards performing \emph{variable-to-constant} comparisons,  
and then show that this equivalence relation is a cost-based clustering.
By focusing on such guards, one can improve performance of alignment-based analytic tasks. 
Indeed, variable-to-constant guards, although simple, are extensively used in practice, and they have been subject to an extensive body of research~\cite{LeoniFM18}. 
Moreover, these guards are common in benchmarks from the literature\todo{For Marco: do we want to say that these are the only ones discoverable?}. 
Note, however, that we do not restrict the DPNs we consider to use only such guards. 

Recalling that constraints are used in DPNs as guards associated to transitions, and that a constraints is in general a boolean expression whose atoms are comparisons (cf. Section~\ref{sec:dpn}), we use $\atom(\constraint)$ to define the set of all atoms in a guard $\constraint\in G_\NN$. 
Given a DPN $\NN$, a \emph{variable-to-constant} atom is an expression of the form 
$x \odot k$, where $\odot \in \set{{>}, {\geq}, {=}}$, $x\in V^r\cup V^w$ and $k$ is a constant in $\mathbb Z$ or $\mathbb Q$\todo{other sorts?}. 
We say that a variable $v\in V$ is \emph{restricted to constant comparison} when all atoms in the guards of $\NN$ that involve $v^r$ or $v^w$ are variable-to-constant atoms. 
For such variables,
we also introduce the set $\mathit{ats}_v = \set{ v\odot k \mid x \odot k\in \atom(\constraint), \text{ for some } \constraint\in G_\NN, x\in \set{v^r, v^w}}$, i.e., the set of comparison atoms $v \odot k$ as above with non-annotated variables. $\mathit{ats}_v$ can be seen as a set of predicates with free variable $v$.
%

Intuitively, the optimal alignment of a log trace, given a cost functions as in Remark \ref{rem:distances}, does not depend on the actual variables values specified in the events in the log trace, but only on whether the atoms in $\mathit{ats}_v$ are satisfied.
In this sense, our approach can be considered as a special form of \emph{predicate abstraction}.
Based on this idea, trace equivalence is defined as follows:

\begin{definition}
\label{def:assignment equivalence}
For a variable $v$ that is restricted to constant comparison and two values $u_1$, $u_2$, 
let $u_1 \equivcc^v u_2$ if for all $v\odot k \in \mathit{ats}_v$, $u_1 \odot k$ holds iff 
$u_2 \odot k$ holds.
Two event variable assignments $\alpha$ and $\alpha'$ are 
\emph{equivalent up to constant comparison}, denoted $\alpha \equivcc \alpha'$, if
$\dom(\alpha)=\dom(\alpha')$ and
for all variables $v \in \dom(\alpha)$, one of the following conditions must hold:
	\begin{compactitem}
		\item $\alpha(v) = \alpha'(v)$; or
		\item $v$ is restricted to constant comparison and $\alpha(v)\equivcc^v \alpha'(v)$.
	\end{compactitem} 
\end{definition}
This definition intuitively guarantees that $\alpha$ and $\alpha'$ ``agree on satisfying'' the same atomic constraints in the process. For example, if $\alpha(x)=4$ and $\alpha'(x)=5$, then, given two constraints $x>3$ and $ x<2$, we will get that $\alpha\models x>3$ and $\alpha'\models x>3$, whereas $\alpha\not\models x<2$ as well as $\alpha'\not\models x<2$.

\begin{definition}[Equivalence up to constant comparison]
\label{def:trace equivalence}
Two events $e = (b, \alpha)$ and $e' = (b', \alpha')$ are 
\emph{equivalent up to constant comparison}, denoted $e\equivcc e'$,
if $b=b'$ and $\alpha \equivcc \alpha'$.

Two log traces $\tau$, $\tau'$ are \emph{equivalent up to constant comparison}, denoted $\tau\equivcc\tau'$,
iff their events are pairwise equivalent up to constant comparison.
That is,
$\tau = \seq\tau$, $\tau' = \seq{\tau'}$, and $\logtrace_i\equivcc\logtrace_i'$
for all $i$, $1\,{\leq}\,i\,{\leq}\,n$.
\end{definition}

\begin{example}
In Ex. \ref{exa:1}, the variable $x$ is restricted to constant comparison, while $y$ is not.
Since $\mathit{ats}_x = \{x \geq 0, x \leq 3\}$,
the log traces 
$\logtrace_1 = \langle (\mathsf{a}, \{x\mapsto 2\}), (\mathsf{b}, \{y\mapsto 1\})\rangle$ and
$\logtrace_2 = \langle (\mathsf{a}, \{x\mapsto 3\}), (\mathsf{b}, \{y\mapsto 1\})\rangle$,
satisfy $\logtrace_1 \equivcc \logtrace_2$, but for
$\logtrace_3 = \langle (\mathsf{a}, \{x\mapsto 4\}), (\mathsf{b}, \{y\mapsto 1\})\rangle$ 
we have $\logtrace_1 \not\equivcc \logtrace_3$ because $3 \not\equivcc^x 4$, and
$\logtrace_4 = \langle (\mathsf{a}, \{x\mapsto 3\}), (\mathsf{b}, \{y\mapsto 2\})\rangle$
satisfies $\logtrace_1 \not\equivcc \logtrace_4$ because the values for $y$ differ.
The equivalent traces $\logtrace_1$ and $\logtrace_1$ have the same optimal cost: 
for the alignments
\[
\gamma_1=
\syncmove{\m a}{x=2}{\m a}{x^w=2}
\syncmove{\m b}{y=1}{\m b}{y^w=1}
\modelmovenw{\tau}
\qquad
\gamma_2=
\syncmove{\m a}{x=3}{\m a}{x^w=3}
\syncmove{\m b}{y=1}{\m b}{y^w=1}
\modelmovenw{\tau}
\qquad
\gamma_3=
\syncmove{\m a}{x=4}{\m a}{x^w=3}
\syncmove{\m b}{y=1}{\m b}{y^w=1}
\modelmovenw{\tau}
\]
we have
$\kappa_\NN^{opt}(\logtrace_1) = \kappa(\gamma_1) = 0$ and
$\kappa_\NN^{opt}(\logtrace_1) = \kappa(\gamma_1) = 0$.
Note, however, that the respective process runs $\restr{\gamma_1}M$ and $\restr{\gamma_2}M$
differ.
On the other hand, $\gamma_3$ is an optimal alignment for $\logtrace_3$ but
$\kappa(\gamma_3) = \kappa_\NN^{opt}(\logtrace_3) = 1$.

Moreover, $\logtrace_1$ and $\logtrace_3$ illustrate that for trace equivalence
it does not suffice to consider model transitions with activity labels that occur in the traces: all events in $\logtrace_1$ and $\logtrace_3$ correctly correspond to transitions
with the same labels in $\NN$, but for a \emph{later} transition the value of $x$
makes a difference. This motivates the requirement that in equivalent traces (Defs. \ref{def:assignment equivalence} and \ref{def:trace equivalence})
the values of a variable $v$ that is restricted to constant comparison satisfies the same
subset of $\mathit{ats}_v$.
\todo[inline]{better explanation welcome (or cut if too verbose)}
\end{example}

We next show that equivalence up to constant comparison is a cost-based 
clustering, provided that the cost function is of a certain format.  
To that end, we consider a distance-based cost function $\kappa$ from Definition~\ref{def:distance} and call it \emph{comparison-based}, when the following conditions hold:
\begin{compactenum}
	\item $P_L(b,\alpha)$ does not depend on the values assigned by $\alpha$, and
	      $P_M(t,\beta)$ does not depend on the values assigned by $\beta$;
	\item the value of $P_=((b,\alpha), (t,\beta))$ depends only on whether conditions $b=\ell(t)$ and $\alpha(v) = \beta(v^w)$ are satisfied or not. 
\end{compactenum}
Note that this requirement is satisfied by the distance-based cost function in Remark \ref{rem:distances}. Indeed, in the standard cost function, $P_L(b,\alpha)=1$ and thus it does not depend on $\alpha$. Moreover, the second condition is clearly satisfied, as in $P_=((b, \alpha), (t,\beta)) = |\{v\in \dom(\alpha) \mid \alpha(v) \neq \beta(v^w)\}|$, for $b = \ell(t)$, we only need to check whether $\alpha(v) \neq \beta(v^w)$.

\begin{theorem}
\label{thm:cluster}
Equivalence up to constant comparison is a cost-based clustering with
respect to any comparison-based cost function.
\end{theorem}
\begin{proof}
We need to show that for any two traces $\logtrace_1$ and $\logtrace_2$ such that $\logtrace_1\equivcc \logtrace_2$ and a comparison-based cost function $\kappa$,
it holds that $\logtrace_1\equiv_{\kappa_\NN^{opt}}\logtrace_1$.
For a partial process run $\sigma$, let $\alpha_{sv}(\sigma)$ be the state variable assignment after the last transition firing of the partial process run $\sigma$.
Note that since $\logtrace_1\equivcc \logtrace_2$, the lengths of the two traces as well as
their sequences of executed activities coincide.
To prove the claim, we verify that if $\logtrace_1$ has an alignment $\gamma_1$ with cost 
$\kappa(\gamma_1) = \delta(\logtrace_1, \procrun_1)$ for some process run 
$\procrun_1=\restr{\gamma_1}{M}$, 
then there is a process run $\procrun_2$ such that 
$\delta(\logtrace_2, \procrun_2) = \kappa(\gamma_1)$, and hence there is an alignment 
$\gamma_2$ with $\restr{\gamma_2}L = \logtrace_2$, $\restr{\gamma_2}M = \procrun_2$
and $\kappa(\gamma_2) = \delta(\logtrace_2, \procrun_2)$.
More precisely, let $|\logtrace_1|=|\logtrace_2|=m$, $\procrun_1=\restr{\gamma_1}{M}$ and 
$|\procrun_1| = n$. Then, we show by induction on $m+n$ that there exists a process run $\procrun_2$ such that $|\procrun_2|=n$, 
$\delta(\logtrace_1,\procrun_1)=\delta(\logtrace_2,\procrun_2)$, 
and $\alpha_{sv}(\procrun_1) \equivcc \alpha_{sv}(\procrun_2)$.
\begin{compactitem}
\item[\textbf{Base case ($m = n = 0$).}] 
In this case all of $\logtrace_1$, $\logtrace_2$, and $\procrun_1$ are empty.
By taking the empty run also for $\procrun_2$, the claim is trivially satisfied as 
$\delta(\epsilon,\epsilon)=0$.
\item[\textbf{Step case ($m > 0$, $n = 0$).}]
By definition, $\delta(\logtrace_1,\epsilon)=P_L((\logtrace_1)_{m}) + \delta(\restr{\logtrace_1}{m-1},\epsilon)$.
As $\logtrace_1\equivcc \logtrace_2$ implies $\restr{\logtrace_1}{m-1}\equivcc \restr{\logtrace_2}{m-1}$,
we can apply the induction hypothesis to obtain
$\delta(\restr{\logtrace_1}{m-1},\epsilon)=\delta(\restr{\logtrace_2}{m},\epsilon)$.
By the assumption $\kappa$ is comparison-based, and activities in $\logtrace_1$ and $ \logtrace_2$ coincide, $P_L((\logtrace_1)_{m}) = P_L((\logtrace_2)_{m})$.
It follows that $\delta(\logtrace_2,\epsilon)=P_L((\logtrace_2)_{m}) + \delta(\restr{\logtrace_2}{m-1},\epsilon)$.
\item[\textbf{Step case ($m = 0$, $n > 0$).}]
Similar as the previous case, using the fact that $P_M((\procrun_1)_{n}) = P_M((\procrun_2)_{n})$ because $\kappa$ is comparison-based.
\item[\textbf{Step case ($m > 0$, $n > 0$).}]
Let $e_1 = (b,\alpha_1) = (\logtrace_1)_{m}$ (resp. $e_2 = (b,\alpha_2) = (\logtrace_2)_{m}$) be
the last event in $\logtrace_1$ (resp. $\logtrace_2$), and
$f = (t, \beta_1)$ the last transition firing in $\procrun_1$.
According to Def. \ref{def:distance}, $\delta(\logtrace_1,\procrun_1)$ is defined
as a minimum of three expressions. Reasoning as in the previous two cases shows that
there are process runs $\hat\procrun_2$, $\overline\procrun_2$ such that
$P_L(e_1) + \delta(\restr{\logtrace_1}{m-1}, \procrun_1) =
P_L(e_2) + \delta(\restr{\logtrace_2}{m-1}, \hat\procrun_2)$ and
$P_M(f) + \delta(\logtrace_1, \restr{\procrun_1}{n-1}) =
P_M((\overline\procrun_2)_{n}) + \delta(\logtrace_2, \restr{\overline\procrun_2}{n-1})$.
We now show that there is also a process run $\procrun_2$ such that
\begin{equation}
\label{eq:delta eq}
P_=(e_1, f) + \delta(\restr{\logtrace_1}{m-1}, \restr{\procrun_1}{n-1}) =
P_=(e_2, (\procrun_2)_{n}) + \delta(\restr{\logtrace_2}{m-1}, \restr{\procrun_2}{n-1})
\end{equation}
so
$\delta(\logtrace_1,\procrun_1)=\delta(\logtrace_2,\procrun_2)$ follows.
As $\logtrace_1\equivcc \logtrace_2$ implies $\restr{\logtrace_1}{m-1}\equivcc \restr{\logtrace_2}{m-1}$,
by the induction hypothesis there exists a process run $\procrun_2'$ such that 
$|\procrun_2'|={n\,{-}\,1}$,
$\delta(\restr{\logtrace_1}{m-1},\restr{\procrun_1}{n-1})=\delta(\restr{\logtrace_2}{m-1},\procrun_2')$, and
$\alpha_{sv}(\restr{\procrun_1}{n-1}) \equivcc \alpha_{sv}(\procrun_2')$.

We set $\procrun_2 = \procrun_2' \cdot (t, \beta_2)$,
 where $\beta_2$ is defined as follows:\footnote{Here, given a process run $\procrun$, its concatenation with a transition firing $f'=(t,\beta)$ is defined as  $\procrun\cdot(t,\beta)=\langle\seq[n] f,f'\rangle$.}
for all $v\,{\in}\,V$, $\beta_2(v^r) = \alpha_{sv}(\procrun_2')(v)$, and $\beta_2(v^w)$ is defined as either
$\beta_2(v^w) =\beta_1(v^w)$ if $v$ is not restricted to constant comparison,
or otherwise
\begin{equation}
\label{eq:beta2}
\beta_2(v^w) =
\begin{cases}
\alpha_2(v) &\text{ if $\beta_1(v^w) = \alpha_1(v)$} \\
\alpha_1(v) &\text{ if $\beta_1(v^w) \neq \alpha_1(v)$ and $\beta_1(v^w) = \alpha_2(v)$} \\
\beta_1(v^w)  &\text{ otherwise}
\end{cases}
\end{equation}
We now show that
\begin{compactenum}[\it (i)]
\item $\beta_2$ satisfies $guard(t)$,
\item $\alpha_{sv}(\procrun_1) \equivcc \alpha_{sv}(\procrun_2)$, and
\item $P_=((b,\alpha_1), (t,\beta_1)) = P_=((b,\alpha_2), (t,\beta_2))$.
\end{compactenum}

For \textit{(i)}, note that 
$\alpha_{sv}(\restr{\procrun_1}{n-1}) \equivcc \alpha_{sv}(\procrun_2')$
implies that for all $v\,{\in}\,V$, either $\beta_1(v^r) = \beta_2(v^r)$, 
or $v$ is restricted to constant comparison and $\beta_1(v^r) \equivcc^v \beta_2(v^r)$.
Moreover, by definition of $\beta_2$ we have for all $v\,{\in}\,V$, either $\beta_1(v^w) = \beta_2(v^w)$, or $v$ is restricted to constant comparison and by Eq. \eqref{eq:beta2}
one of the following holds:
$\beta_2(v^w) = \alpha_2(v) \equivcc^v \alpha_1(v) = \beta_1(v^w)$, or
$\beta_2(v^w) = \alpha_1(v) \equivcc^v \alpha_2(v) = \beta_1(v^w)$, or
$\beta_1(v^w) = \beta_2(v^w)$;
where we use $\alpha_1(v) \equivcc^v \alpha_2(v)$, which follows from $\logtrace_1 \equivcc \logtrace_2$.
Thus, we have the following ($\star$): $\beta_1$ and $\beta_2$ coincide on all variables that are not restricted to constant comparison, and satisfy $\beta_2(v^w) \equivcc^v \beta_1(v^w)$ otherwise. It follows that
since $\equivcc$-equivalent assignments satisfy the same constraints,
and $\beta_1 \models guard(t)$, also $\beta_2 \models guard(t)$.
Item \textit{(ii)} then follows from ($\star$) and the construction of a state variable assignment after a transition firing.

For \textit{(iii)}, we observe that for all variables $v$ such that $\beta_1(v^w) \neq \beta_2(v^w)$, i.e.,
$\beta_2(v^w)$ is defined by one of the three cases in Eq. \eqref{eq:beta2}, one can check that
$\beta_1(x^w) = \alpha_1(x)$ if and only if $\beta_2(x^w) = \alpha_2(x)$.
As $\kappa$ is a comparison-based cost function, it follows that
$P_=((b,\alpha_1), (t,\beta_1)) = P_=((b,\alpha_2), (t,\beta_2))$.

From \textit{(i)} we obtain that $\procrun_2$ is indeed a (partial) process run in $\NN$,
and \textit{(iii)} implies Eq. \eqref{eq:delta eq}.
\qed
\end{compactitem}
\end{proof}

\noindent
The constructive proof of Thm. \ref{thm:cluster}  shows that given an optimal alignment $\gamma$ for a log trace $\logtrace$, an optimal alignment for any equivalent trace $\logtrace'$ (so that $\logtrace \equivcc \logtrace'$) is easily computed from $\gamma$, $\logtrace$, and $\logtrace'$ in linear time.

\section{Implementation and Experiments}
\label{sec:implementation}

As a proof of concept, we implemented the DPN conformance checking tool
\tool based on the encoding in Sec. \ref{ssec:encoding}. 
In this section we comment on its implementation, some optimizations,
and experiments on benchmarks from the literature. The source code is publicly available.\footnote{\url{https://github.com/bytekid/cocomot}}

\paragraph{Implementation.}
Our \tool prototype is a Python command line script:
it takes as input a DPN (as \texttt{.pnml} file) and a log (as \texttt{.xes}) and computes the optimal alignment distance for every
trace in the log. In verbose mode, it additionally prints an
optimal alignment.
To reduce effort, \tool first preprocesses the log to a sublog of unique traces, and second applies trace clustering as described in Sec. \ref{sec:clustering} to further partition the sublog into 
equivalent traces.
The conformance check is then run for one representative from every 
equivalence class.

The tool \tool uses \texttt{pm4py}~\cite{pm4py} to parse traces,
and employs the SMT solver Yices 2~\cite{Dutertre14} , or alternatively Z3~\cite{deMouraB08}, as backend.
Instead of writing the formulas to files, we use the bindings
provided by the respective Python interfaces~%
\cite{yicesinterface,z3interface}.
Since Yices 2 has no optimization built-in, we implemented a
minimization scheme using multiple satisfiability checks.
Every satisfiability check is run with a timeout, to avoid
divergence on large problems.

\paragraph{Encoding optimizations.}
To prune the search space, we modified the encoding presented in Sec. \ref{ssec:encoding}. 
The most effective changes are the following ones:
\begin{itemize}
 \item 
 We perform a reachability analysis in a preprocessing step. This allows us to restrict
 the range of transition variables $t_i$ in \eqref{eq:transition range}, as well as
 the cases $t_i=j$ in \eqref{eq:enabled} and \eqref{eq:token game} to those that are actually reachable.
 Moreover, if a data variable $v\in V$ will never be
 written in some step $i$, $1\leq i \leq n$, because no respective transition is 
 reachable, we set $X_{i,v}$ identical to $X_{i-1,v}$ to reduce the number of
 variables.
\item
If the net is 1-bounded, the marking variables $M_{i,p}$ are chosen boolean rather than integer, similar as in~\cite{BoltenhagenCC19}.
\item
As $\delta_{m,n}$ is minimized, the equation of the form
$\delta_{i+1,j+1} = min(e_1,e_2,e_3)$ 
in ($\phi_\delta$) can be replaced by inequalities
$\delta_{i+1,j+1} \geq min(e_1,e_2,e_3)$.
The latter is equivalent to 
$\delta_{i+1,j+1} \geq e_1 \vee \delta_{i+1,j+1} \geq e_2\vee \delta_{i+1,j+1} \geq e_3$,
which is processed by the solver much more efficiently
since it avoids an if-then-else construct.
\item Several subexpressions were replaced by fresh variables
(in particular when occurring repeatedly),
which had a positive influence on performance.
\end{itemize}

\paragraph{Experiments.}

We tested \tool on three data
sets also used in earlier work~\cite{Mannhardt18,MannhardtLRA16}.
All experiments were run single-threaded on a 12-core
Intel i7-5930K 3.50GHz machine with 32GB of main memory.
\begin{itemize}
 \item The \emph{road fines} data set contains 150370 traces (35681 unique) of
 road fines issued by the Italian police.
 By trace clustering the log reduces to 4290 non-equivalent traces. In 268 seconds, \tool computes optimal alignments for all traces in this set; spending 13\% of the computation time on parsing the log, 13\% on the generation of the encoding, and the rest in the SMT solver.
 When omitting the clustering precprocessor, \tool requires about 30 minutes to process the 35681 traces.
 We note some data about the model and log:
 The maximal length of a trace is 20, and its average alignment cost is 1.5. The average time spent on a trace is 0.1 seconds. The process model has less than 20 transitions, and at most one token around at any point in time.
 \item 
 The \emph{hospital billing} log contains 100000 traces (4047 unique) of a hospital billing process. Trace clustering slightly reduces the number of non-equivalent traces to 4039. For 3392 traces \tool finds an optimal alignment, while SMT timeouts occur for the remaining, very long traces (the maximal trace length is 217).
 \item 
 The \emph{sepsis} log contains 1050 unique (and non-equivalent) traces.
 For 1006 traces \tool finds an optimal alignment, while it times out for the remaining, very long traces (the maximal trace length is 185).
\end{itemize}

For the experiments described in above we used Yices (with an SMT timeout of 10 minutes)
since Z3 turned out to be considerably slower: checking conformance of the road fine log using Z3 (with its built-in minimization routine) takes more than two hours.
It is notable that across all data sets, only 1\% of the computation time is spent on generating the encoding, while the vast majority of the time is used for SMT solving.

\section{Discussion}
\label{sec:discussion}

In this section we outline how the \cocomot approach, due to its modularity, readily 
lends itself to further tasks related to the analysis of data-aware processes.

\smallskip
\noindent
 The \textbf{multi-alignment} problem asks, given a DPN $\NN$ and a set of log traces 
$\{\seq[n]\logtrace\}$, to find a process run $\procrun \in \PP_\NN$ such that 
$\sum_{i=1}^n \kappa(\gamma_i)$ is minimal, where $\gamma_i$ is a minimal-cost alignment of $\logtrace_i$ and $\procrun$ for all $i$, $1\leq i \leq n$~\cite{ChatainCD17}.\footnote{Instead of the sum, also other aggregation functions can be used, e.g., maximum.}
Our encoding can solve such problems by combining $n$ copies of the
distance variables and their defining equations $(\varphi_\delta)$ with \eqref{eq:init}--\eqref{eq:data}, and minimizing the above objective.
Generalizing alignments,
multi-alignments are of interest for their own sake, but also
useful for further tasks, described next.

\smallskip
\noindent
 \textbf{Anti-alignments} were introduced to find
model runs that deviate as much as possible from a log, e.g. for precision checking~\cite{ChatainC16}. For a set of traces
$\{\seq[n]\logtrace\}$, the aim is to find $\procrun \in \PP_\NN$ of bounded length such that 
$\sum_{i=1}^n \kappa(\gamma_i)$ is \emph{maximal}, with $\gamma_i$ is as before.
Using our encoding, this can be done as in the multi-alignment case, replacing minimization by maximization.

\smallskip
\noindent 
\textbf{Trace clustering} was studied as a method to partition event logs into more homogeneous sub-logs, with the hope that process discovery techniques will perform better on the sub-logs than if applied to the original log~\cite{Aalst11,ChatainCD17}.
Chatain \textsl{et al}~\cite{ChatainCD17,BoltenhagenCC19} propose trace clustering based on multi-alignments. In the same fashion, our approach can be used to partition a log of DPN traces.

\smallskip
\noindent
Our approach can also be used for \textbf{model repair} tasks:
given a set of traces, we can use multi-alignments to minimize the sum of the trace distances, while replacing a
parameter of the DPN by a variable (e.g., the threshold value in a guard).
From the satisfying assignment we obtain the value for this parameter
that fits the observed behavior best.
As constraints \eqref{eq:init}--\eqref{eq:data} symbolically describe a process run of bounded length,
our encoding supports \textbf{bounded model checking}. Thus we could also implement 
\emph{scenario-based} conformance checking, to find for a given trace the best-matching process run that satisfies additional constraints, e.g., that certain data values are not exceeded.

\medskip
\noindent 
Finally but crucially, the \textbf{main advantage of SMT} 
is that it offers numerous \emph{background theories} to capture the data manipulated by the DPN, and to express sophisticated cost functions. The approach by Mannhardt \textsl{et al} ~\cite{Mannhardt18,MannhardtLRA16} needs to restrict guards of DPNs to linear arithmetic expressions in order to use the MILF backend. 
In our approach, the language of guards may employ \emph{arbitrary} functions and predicates from first-order theories supported by SMT solvers (e.g., uninterpreted functions, arrays, lists, and sets). For example, the use of relational predicates would allow to model structured background information, and possibly even refer to full-fledged relational databases from which data injected in the net are taken. Moreover, the background theory allows to express sophisticated cost functions, as in Def. \ref{def:distance} with the following parameters (inspired by~\cite{Mannhardt18}):
$P_=((b, \alpha), (t,\beta)) = |\{v\in write(t) \mid \neg R(\alpha(v)), R( \beta(v^w)) \}|$ if $b = \ell(t)$, for some relation $R$ from a database $DB$: in this way, $P_=$ counts the number of written variables whose values in the model run are stored in the relation $R$ from $DB$ whereas their values in the log trace are not.

\section{Conclusions}
We have introduced \cocomot, a foundational framework equipped with a proof-of-concept, feasible implementation for alignment-based conformance checking of multi-perspective processes. Beside the several technical results provided in the paper, the key, general contribution provided by \cocomot is to connect the area of (multi-perspective) conformance checking with that of declarative problem solving via SMT. This comes with a great potential for homogeneously tackling a plethora of related problems in a single framework with a solid theoretical basis and several state-of-the-art algorithmic techniques, as shown in Sec.~\ref{sec:discussion}. 
The support of databases, as well as the use of complex SMT features for expressive cost functions, are left for future work, but motivate once again the use of SMT.

\bibliographystyle{abbrv}
\bibliography{references}
\end{document}